\documentclass[twoside]{article}

\usepackage[accepted]{aistats2023}
\usepackage{amssymb,mathrsfs}
\usepackage{amsmath}
\usepackage{amsthm}

\usepackage[linesnumbered, vlined, ruled]{algorithm2e}
\usepackage{graphicx}
\usepackage{multirow}
\usepackage[hidelinks]{hyperref}
\usepackage[round]{natbib}
\usepackage{threeparttable}

\usepackage{color}
\usepackage{soul}
\usepackage{cancel}

\begin{document}
%

%

\twocolumn[

\aistatstitle{GBOSE: Generalized Bandit Orthogonalized Semiparametric Estimation}

\aistatsauthor{ Mubarrat Chowdhury$^*$ \And Elkhan Ismayilzada$^*$ \And Khalequzzaman Sayem$^*$ \And Gi-Soo Kim$^\dagger$ }

\aistatsaddress{ UNIST \And  UNIST \And UNIST \And UNIST } ]

\begin{abstract}
    
    In sequential decision-making scenarios i.e. mobile health, recommendation systems, revenue management; contextual multi-armed bandit algorithms have garnered attention for their performance. But most of the existing algorithms are built on the assumption of a strictly parametric reward model, mostly linear in nature. In this work, we propose a new algorithm with a semi-parametric reward model with state-of-the-art complexity of upper bound on regret amongst existing semi-parametric algorithms. Our work expands the scope of another representative algorithm \citep{krishnamurthy2018semiparametric} of state-of-the-art complexity with a similar reward model by proposing an algorithm built upon the same action filtering procedures but provides explicit action selection distribution for scenarios involving more than two arms at a particular time step while requiring fewer computations. We derive the said complexity of the upper bound on regret  and present simulation results that affirm our method’s superiority out of all prevalent semi-parametric bandit algorithms for cases involving over two arms.

\end{abstract}
\section{Introduction}
\def\thefootnote{*}\footnotetext{Equal contribution.}
\def\thefootnote{$\dagger$}\footnotetext{Corresponding author.}
Sequential decision-making has enormous influences on our day-to-day activities. Starting from news or movie recommendations to clinical trials, the previous decision history plays a significant role to make the next decision. For instance, in news recommendation services, the server also referred to as the learner or agent, suggests a news article to the user based on the user's previous records such as search history, preferences, demographics, etc with the goal of maximizing the engagements of the user, which in a bandit setting can be deemed as the reward. In most scenarios, the users' history is not previously gathered. Hence, to optimize decision-making, a proper procedure is obligatory for controlling the pursuit of acquiring the user preference information by taking sub-optimal decisions, i.e., exploration in a bandit setting, and the use of collected information to make better decisions, i.e., exploitation in a bandit setting. 

The solution to the aforementioned problem can be described using the multi-armed bandit framework \citep{lai1985asymptotically}. The extended version of traditional multi-armed bandit (MAB), contextual bandit, that additionally considers the context information in the decision process, has shown promise in different fields, such as mobile healthcare system \citep{tewari2017ads}, ad \citep{langford2008exploration} and news recommendations \citep{kawale2015efficient}, revenue management \citep{ferreira2018online}, etc. In Contextual bandit scenario, at each time step, a learner aims to select the option/arm/action expected to yield the highest reward. Upon selection, the learner receives a reward yielded by the environment, the mean value of which is unknown but depends on the context under contextual bandit settings. Regret can be defined as the difference between the reward that would have been yielded from the optimal action and the actual reward yielded by the chosen action. Bandit algorithms seek to minimize cumulative regret across time.

Based on their reward and contextual information relationship assumptions contextual bandits can primarily be classified into two categories, parametric and agnostic. The majority of the parametric algorithms assume a linear reward model \citep{filippi2010parametric,abbasi2011improved,chu2011contextual,agrawal2013thompson}. Upon the linear model function learned with high accuracy, the algorithms can take near-optimal decisions at each time step and onward. However, in the real world, rarely any reward model is linear; which may result in the unsatisfactory performance of these algorithms in practice, since the rather simplistic primary assumption may not hold.  

Contrary to the parametric models, as the name suggests model agnostic contextual bandit settings do not assume a predefined policy \citep{langford2008exploration, agarwal2014taming}. From the oracle of policies that are parameterized in some way, the algorithms choose the best-fitted policy for the given data.
While model-agnostic algorithms are able to describe non-linear scenarios, they are typically computationally intensive, fall behind the best statistical guarantees compared to their parametric counterpart, and might require solving NP-hard problems in worst-case scenarios.

To address the issue, a few of the recent works introduced a new setup, semi-parametric contextual bandit where the reward is modeled as a linear function of the decision confounded by an additive non-linear perturbation that is independent of decision \citep{greenewald2017action, krishnamurthy2018semiparametric, kim2019contextual}. The confounding term significantly generalizes the linear parametric setting, enabling them to handle more complex, non-stationary, and non-linear reward models. Though \cite{greenewald2017action} have considered the non-stationary assumption, the performance of the method is guaranteed under a restrictive condition on action choice probabilities. While \cite{kim2019contextual} proposed a method without any restriction and computationally efficient than the methods of \cite{greenewald2017action} and \cite{ krishnamurthy2018semiparametric}, the BOSE algorithm by \cite{krishnamurthy2018semiparametric} currently holds the state of the art complexity for upper regret bound for the semi-parametric contextual bandit problems. However, in their work, the distribution for action selection is not specified explicitly for the cases when the number of arms/actions, $N > 2$. This could make the algorithm impractical in real-world scenarios which transpire to have more than two arms in most of the cases.  In our work, we propose a new algorithm Generalized Bandit Orthogonalized Semiparametric Estimation (GBOSE) that expands the scope of BOSE by proposing an algorithm that is built upon the same action filtering procedures but a new action selection distribution which is expandable to scenarios involving more than two arms at a particular time step while requiring fewer computations. We present the proof for the complexity of the upper bound on cumulative regret that shows that the state-of-the-art (SOTA) $\tilde{\mathcal{O}}(d\sqrt{T})$ regret bound is preserved in our proposed algorithm, where $\tilde{\mathcal{O}}()$ ignores the logarithmic terms. After the proof, we present simulation study results that show our method leads with the least cumulative regret among other existing semi-parametric methods \citep{greenewald2017action, kim2019contextual} and Thompson Sampling for linear reward model \citep{agrawal2013thompson}. We are second best to Thompson Sampling, an algorithm with a linear model assumption of reward,  only when the simulation setting is purely linear. However, real-world scenarios are believed to be non-linear and the resilience of our algorithm in non-linear cases makes it more desirable in practice. 
\section{Preliminaries}
In bandit scenarios, for each time $t$,  a learner chooses an action or arm $i$ from $N$ number of actions or arms, where each arm is assumed to have a finite-dimensional context vector $b_{t, i} \in \mathbb{R}^d$. For each action, the learner receives a random reward $r_{t, i}$, which is of unknown mean $\mu$ but depends on the context vector $b_{t, i}$, with an arbitrary function $\theta (t) = \theta_t(b_{t, i})$. After choosing an arm $a_t$ out of $N$ arms at time $t$, the reward for that time-step $r_{t, a_t}$ is observed. The optimal arm $a_t^*$ at time step $t$ is the arm that maximizes the reward, i.e., $a_t^* := \underset{1 \leq i \leq N}{\text{argmax}}$ $ \theta_t(b_{t, i})$. The difference between the reward of the optimal arm and the chosen arm, namely the regret, can be written as follows.

 \begin{equation}
    \begin{aligned}
    \text{regret($t$)} &=  \mathbb{E}\left(r_{t,a_t^*} - r_{t,a_t} | \{b_{t, i}\}_{i=1}^{N}, a_t\right)\\
    &= \theta_t(b_{t, a_t^*}) - \theta_t(b_{t, a_t})& 
    \end{aligned}\label{eq1}
\end{equation}

Bandit algorithms aim to minimize the cumulative regret, $R(T) = \sum_{t = 1}^T \text{regret($t$)}$ over the time period $T$. For the linear parametric algorithms, the reward is formulated as follows with unknown $\mu \in \mathbb{R}^d$,
$$r_{t,i} = \theta_t(b_{t,i}) + \varepsilon_{t, i} = b_{t, i}^T \mu + \varepsilon_{t, i}, i=1, \dots, N$$
In this work, where we study the generalization of the linear stochastic bandit with a confounder function $v(t)$ added to the action-dependent features, the reward received by the learner is
\begin{equation}
    r_{t,i} = \theta_t(b_{t,i}) + v(t) + \varepsilon_{t, i} = b_{t, i}^T \mu + v(t) + \varepsilon_{t, i}, i=1, \dots, N
    \label{eq2}
\end{equation}
Error term $\varepsilon$ is an i.i.d. noise process in both parametric and our more relaxed semi-parametric settings' reward received by the learner.
Given the history $\mathcal{H}_{t-1} = \{ a_\mathcal{T}, r_{a_\mathcal{T}}, \{b_{\mathcal{T}, i}\}_{i=1}^{N}, \mathcal{T} = 1 , \dots , t - 1 \} $ of reward and arm chosen by the learner until $t-1$ and filtration $\mathcal{F}_{t-1}$ which is the union of $\mathcal{H}_{t-1}$ and the context at time $t$, $\{b_{t, i}\}_{i=1}^{N}$, the expected reward is:
$$ E(r_{t, i} | \mathcal{F}_{t-1}) = b_{t, i}^T \mu + v(t)$$
Although we have a non-linear term $v(t)$ added to our expected reward, we seek to optimize the same regret shown in equation (\ref{eq1}), wherein the final expression, $v(t)$ is not included since it is independent of action choice. To formulate such scenarios, we have two assumptions to proceed, \\
\textbf{Assumption 1:} We assume that at each time $t$, context vector $\{b_{t, i}\}_{i=1}^{N}$, confounder function $v(\cdot)$ and the i.i.d noise $\varepsilon$ are determined by the environment given $\mathcal{H}_{t-1}$ before arm $a_t$ is chosen.  An adaptive adversary provides $\{b_{t, i}\}_{i=1}^{N}$ and $v(\cdot)$ that satisfies $\mathbb{E}\left(\varepsilon_t | \{b_{t, i}\}_{i=1}^{N}, v(t)\right) = 0$ and $|\varepsilon_{t}| \leq 1$. \\
\textbf{Assumption 2:} $|| \mu ||_2$  and  $||b_{t,i}||_2$,  $\forall i \in [N]$, are assumed to be less than or equal to 1 and $v(t)$ is bounded between -1 to 1. Here, $|| \cdot ||_2$ means the $L_2$ norm.\\
Despite our algorithm's adaptability for more general bounds, for the sake of simplicity of analysis, we restricted the bound to -1 to 1 for $v(t)$ and the upper bound for $||\mu||_2$ and $||b||_2$ to 1.
\section{Related Works}
Semi-parametric bandit has been discussed in several studies in these years. \cite{greenewald2017action} first introduced a setting similar to ours, introduced a new reward estimator for Thompson Sampling \citep{agrawal2013thompson} that is consistent despite the confounding term. Because of having medical applications prioritized for their work, they used the first action for the base action as the assumption and set the context vector to zero and the reward for the first arm is calculated based on the confounder function's output. Among the two steps of their algorithm, in the first step, they select an action from the remaining non-base actions using a similar approach as Thompson sampling. Thenceforward they choose the final action arm between the base action and the arm selected from in the previous step with a probability $ \in [p, 1 - p]$ for some small $p \in (0, 1)$. The authors introduced a new regret notion that is different than ours and the results are not exactly equivalent to ours. If translated to our regret notion, it produces $O(Md^\frac{3}{2}\sqrt{T}\sqrt{log(T/\delta)^3})$ upper regret bound which is worse than ours $O(d\sqrt{T}log(T/\delta))$. Here, $\delta \in (0, 1)$; $M$ is a constant, dependent on $p$ and $M$ becomes an extremely large number if less restriction is enforced on the arm selection, wherein in ours, we do not set any restriction in arm selection. 

\cite{kim2019contextual}, who also consider the same setting as ours, improved the method of \cite{greenewald2017action} by providing a method without any restriction. They introduced a new estimator for the Thompson Sampling framework, that enables them to have tighter regret bound than \cite{greenewald2017action}. The method is shown to have $O(NT)$ time complexity as it derives a probability distribution over the arms in $O(N)$ computations, yet the upper bound of regret decreases to $O(d^\frac{3}{2}\sqrt{T}\sqrt{log(T/\delta)^3})$ from the regret upper bound of \cite{greenewald2017action}. While their method has lower time complexity compared to ours, $O(N^2T)$, our method ensures lower and current state-of-the-art upper regret bound. 

Our work has the closest relationship with \cite{krishnamurthy2018semiparametric}, which currently provides the least upper regret bound for our setting. Adapted from \cite{even2006action} given policy of arm elimination, they procured an action elimination step to filter out arms before choosing the final arm in the second step. They as well introduced a new distribution to select the final arm from which incur $O(d\sqrt{T}log(T/\delta))$ upper regret bound, current best for the setting. However, the given distribution is given explicitly  if the number of arms is only two. In our method, while using the same filtration method to eliminate arms, we specify an explicit  technique for finding a distribution that is proven to lead to an upper bound of regret of the same order, $O(d\sqrt{T}log(T/\delta))$, for a high probability $ 1- \delta$. In addition to that, our method has less computational overhead compared to \cite{krishnamurthy2018semiparametric}. More details and proofs are described in Section \ref{algorithm} and \ref{proof}.  

Our setting may also be possible to describe using model agnostic contextual bandit approaches such as adversarial bandits where the reward is estimated from a class of large {$k$} parameterized models, without any presumption \citep{auer2002using,langford2008exploration,beygelzimer2011contextual, agarwal2014taming}. But finding the optimal policy for the given reward history increases the computational overhead extremely high as often policy has to be searched in infinitely large {$k$} number of policy classes. It might require solving the NP-hard optimization problem. More importantly, unlike ours, the regret bound expands with $\sqrt{k}$ for most of them. Since we have a  prior assumption of a parametric model and add a confounder term to make it non-linear in our setting, the number of computations decreases while ensuring performance.  

 Most of the other bandit algorithms primarily have an assumption of a linear reward model, which may not be strictly comparable to our setting \citep{li2010contextual,filippi2010parametric, chu2011contextual, abbasi2011improved, agrawal2013thompson}. In fact, except for Agrawal and Goyal (2010), which uses Bayesian inferences to calculate reward, the majority of them are deterministic algorithms. Although they probably will fail and could incur $\Omega (T)$ regret in our setting since the underlying model is different, verifying the fact is extremely challenging. In the experiments, we verify the performance of LinTS (\cite{agrawal2013thompson}) which assumes a linear reward model. LinTS shows robust performance even under nonstationary reward models and sometimes outperforms existing semiparametric algorithms.   
\section{Proposed Method}
This section presents the proposed algorithm for the semi-parametric reward setting(\ref{eq2}). The proposed algorithm expands the applicability of the BOSE algorithm \citep{krishnamurthy2018semiparametric} by specifying distributions for arms that involve more than two cases, while still preserving the SOTA complexity on the upper bound of regret which is lowest among algorithms that consider similar reward setting. This section also includes the theoretical result presenting the preserved $\Tilde{\mathcal{O}}(d\sqrt{T})$ regret bound for the proposed algorithm. 

\subsection{Algorithm} \label{algorithm}

\begin{algorithm*}[!ht]
\textbf{Input}: $T,\delta\in\left(0,1\right)$\\
\textbf{Set} $\lambda \leftarrow 4d \log(9t)+8log(4T/\delta),\gamma(T)\leftarrow \sqrt{\lambda} + \sqrt{27d\log(1+2T/d) + 54 \log(4T/\delta)})$ \\
 \textbf{Initialize} $\hat{\mu}_{0}\gets0\in\mathbb{R}^d,B_{0}\in\lambda I_{d\times d}$ \\
\For{$t=1,\ldots,\ T$}
{
 Observe $b_{t,a}$ \\
 Filter:
\begin{equation}
   \mathcal{A}_{t} \in \{ i \in [N]| \forall j \in [N],
\langle \hat{\mu}_{t-1},b_{t,j} - b_{t,i}\leq \gamma(T)||b_{t,i}-b_{t,j}||_{B_{t-1}^{-1}} \rangle\}
\end{equation} \\
 Set $k,l = \underset{i, j \in \mathcal{A}_t}{argmax} ||b_{t,j} - b_{t,i}||_{B_{t-1}^{-1}}$ \\
 Set $\pi_{k}=\frac{1}{2}$, $\pi_{l}=\frac{1}{2}$,$\pi_{i}=0$   $\forall$ $i \in \mathcal{A}_{t} $ (where $i \neq k$ or $l$) \\
 Compute $\bar{b}_{t} = \sum_{i \in \mathcal{A}_{t}}\pi_{i}\cdot b_{t,i}$ \\
 Sample $a_{t} ~ \pi_{t}$ and play $a_{t}$. \\
 Observe $r_{t,a_{t}} = \langle\mu, b_{t,a_{t}} \rangle + v(t) + \varepsilon_{t}$ \\
 {Update $B_{t-1}$ and $\hat{\mu}_{t-1}$:
\begin{equation}
    B_{t} \leftarrow B_{t-1} + (b_{t.a_{t}}-\bar{b}_{t})(b_{t.a_{t}}-\bar{b}_{t})^{T},
\end{equation}
\begin{equation}
    \hat{\mu}_{t} \leftarrow B_{t}^{-1}\sum_{\tau = 1}^{t}(b_{\tau,a_{\tau}} - \bar{b}_{\tau})r_{\tau,a_{\tau}}
\end{equation}}
}
\caption{GBOSE: Generalized Bandit Orthogonalized Semiparametric Estimation}
\label{algo}
\end{algorithm*}

Pseudocode for our proposed algorithm GBOSE is presented in {Algorithm \ref{algo}}. Similar to its predecessor,  using the estimate $\hat{\mu}$ of $\mu^{*}$, the algorithm chooses the action in two steps. In the first step, suboptimal arms are filtered out. Then in the sampling step that follows, a probability distribution over the actions that survived the first step is determined and used to sample a single action. The probability distribution as well as the contextual feature vector representing the selected action, and the observed reward are used to update $\hat{\mu}$.

In the bandit algorithms literature, algorithms \citep{filippi2010parametric, abbasi2011improved, chu2011contextual, agrawal2013thompson} that assume a linear relationship between the expectation of reward $r_{t,a_{t}}$ and the context vector of the chosen action $b_{t,a_{t}}$,(expressed as,$E\left(r_{t,a_t}\right)=b_{t,a_t}^T\mu^\ast$), can exploit this property to use the following ridge estimator:
\begin{equation*}
    \hat{\mu}_{Ridge} \triangleq (\lambda I+ \sum_{\tau = 1}^{t}b_{\tau,a_{\tau}}b_{\tau,a_{\tau}}^{T})^{-1}\sum_{\tau=1}^{t}b_{\tau,a_{\tau}}r_{\tau,a_{\tau}}.
\end{equation*}
where $\lambda > 0$.

However, in a setting such as ours, where a non-zero, non-stationary confounding term $v(t)$ is assumed to be added on top of the reward expectation of linear stochastic bandits (expressed as $E\left(r_{a_t}\right)=b_{t,a_t}^T\mu^\ast+v(t)$),  the estimator $\hat{\mu}_{Ridge}$ no longer remains a reliable estimator of $\mu^*$.\\

\cite{krishnamurthy2018semiparametric} get around this problem by centering the context vector of the selected action $b_{a,t}$ with mean feature vector of surviving actions of first filtering step, $\bar{b_{t}}$. After the first selection step, for all the arms in the surviving set $\mathcal{A}$ a distribution $\pi_{t} \in \bigtriangleup(\mathcal{A})$ is determined. Therefore, $\bar{b_{t}} = \mathbb{E}_{a_{t}\sim\pi_{t}}[b_{t,a_{t}}|\mathcal{F}_{t-1}]$. This centering operation is known in the literature as \textit{Neyman Orthogonalization} \citep{neyman1979c}, analyzed by \cite{robinson1988root} under the context of linear regression, by \cite{chernozhukov2016double} under a more generalized setting, by \cite{kim2019contextual} under the bandit setting besides \cite{krishnamurthy2018semiparametric}.

Hence for our setting, in the case of BOSE and GBOSE, the orthogonalized estimator can be expressed as:

\begin{flalign*}
    B_{t} = \lambda I + \sum_{\tau = 1}^{t}(b_{\tau,a_{\tau}}-\bar{b_{\tau}})(b_{\tau,a_{\tau}}-\bar{b_{\tau}})^{T}\\
    \hat{\mu}_{t} = B_{t}^{-1}\sum_{\tau = 1}^{t}(b_{\tau,a_{\tau}}-\bar{b_{\tau}})r_{\tau,a_{\tau}}
\end{flalign*}
The $\hat{\mu}$ is the ridge regression version of Robinson's semiparametric regression estimator \citep{robinson1988root}, where the estimator was originally intended for observational studies where the mean of the feature vectors was not known exactly. While attempting to estimate $v(t)$ and the mean value of the feature vectors with which to center the feature vector, if either of the estimates is accurate, the orthogonalized estimator is deemed accurate. In our case, since the distribution $\pi_t$ is determined by the learning agent, the mean value of the feature vector is known to the agent, hence we can use the inaccurate estimate of the confounding term $\hat{v}(t)\equiv0$.\cite{krishnamurthy2018semiparametric} presents an error guarantee for the orthogonalized estimator $\hat{\mu}$ for BOSE by proving a finite concentration inequality for this orthogonalized estimator to show that the confounding term $v(t)$ does not introduce any bias. Since we use this estimator, this proof holds for GBOSE too and is presented in \textbf{Lemma 1}.

\textbf{First action filtration step: } After $b_{t,i}$'s for a time step are observed, both GBOSE and BOSE filter out an action $i$ if it fails to satisfy the condition in \textbf{Algorithm \ref{algo}}. Given the confidence bound in \textbf{Lemma 1}, failure of an arm $i$ to satisfy the condition for a given action $j$ implies a higher expected reward for action $j$, hence arm $i$ is deemed suboptimal with a high probability and dropped.

\textbf{Determining the distribution of surviving actions: } The second stage involves finding a distribution over surviving actions, that satisfy the condition {$||b_{t,i}-\bar{b_{t}}||_{B_{t-1}^{-1}}^{2}\leq4\sum_{j\in \mathcal{A}_t}{\pi_j\left(t\right)||b_{j,i}-\bar{b_{t}}||_{B_{t-1}^{-1}}^2}$, where $\pi_{j}(t)$ refers to the probability assigned to arm $j$}. When the cardinality of the action set $\mathcal{A}_{t}$ of arms that survived the first filtering step is over 2, BOSE just shows, without explicitly specifying the distribution, that an action selection distribution that holds the inequality with no multiplicative constant 4 exists. But constructing such distribution involves computationally intensive operations that require solving a convex program with $N$ quadratic conditions at every iteration. We propose instead a simple distribution that can be computed using the by-products of the first filtering step. We simply assign probability $1/2$ for the pair of arms $j$ and $i$ that yields the highest value for the expression $||b_{t,j}-b_{t,i}||_{B_{t-1}^{-1}}$ and assign a probability 0 to the rest of the arms. We prove that this simple distribution satisfies the constraint of \cite{krishnamurthy2018semiparametric} up to a multiplicative constant 4 which does not affect the complexity of the final regret bound. Note that the pair differences $||b_{t,j}-b_{t,i}||_{B_{t-1}^{-1}}$ are all computed in the first step.
The action selection distribution assignment step in \citep{krishnamurthy2018semiparametric} is not explicitly specified for a problem scenario involving more than 2 arms. \cite{krishnamurthy2018semiparametric} proves that a distribution exists for surviving actions that manage to hold the proven regret bound without explicitly specifying what the distribution is. One of our key contributions is specifying an arm selection distribution that extends to any number of arms. This requires no additional computation besides the ones that have already been performed in the first stage of arm selection. This work derives the regret bound for the proposed GBOSE algorithm which maintains the same order of complexity as BOSE while being easily expandable to problem scenarios for any number of arms.

\subsection{Proof Sketch} \label{proof}
This section lays out the sketch of proof on the complexity of the upper bound on cumulative regret and all the other relevant theorems leading up to that. 
For simplicity, henceforth, we express $\mu^{*}$ as $\mu$.

\textbf{Lemma 1.} \label{lamma_1} Under Assumption 1 and Assumption 2, with probability at least $1-\delta$, for $N=2$ and $\gamma(T) \triangleq \sqrt{\lambda} + \sqrt{9d \log{(1+T/(d\lambda))+18\log{(T/\delta)}}}$ the following bound holds for all $t\in [T]:$
\begin{flalign*}
    ||\hat{\mu}_{t-1} - \mu||_{B_{t-1}} \leq \gamma(T)
\end{flalign*}
Proof. Using {Assumption 1} and definitions, $\hat{\mu}$ can be expressed as:
\begin{flalign*}
    \hat{\mu}_{t-1} = B_{t-1}^{-1}(B_{t-1}-\lambda I)\mu + B_{t-1}^{-1}\sum_{\tau =1 }^{t-1}X_{\tau}\zeta_{\tau}
\end{flalign*}
where $X_{\tau} \triangleq b_{\tau,a_{\tau}} - \bar{b}_{\tau}$ and $\zeta \triangleq  \langle\mu,\bar{b_{\tau}}\rangle + v(\tau) + \varepsilon_{\tau}$ Define $S_{t} \triangleq \sum_{\tau = }^{t-1}X_{\tau}\zeta_{\tau}$. From this we can arrive at:
\begin{flalign*}
    \hat{\mu}_{t-1} - \mu &= B_{t-1}^{-1}(B_{t-1}-\lambda I)\mu +B_{t-1}^{-1}S_{t} - \mu &\\
    &= (\mu - B_{t-1}^{-1}\lambda\mu) +B_{t-1}^{-1}S_{t} - \mu &\\
    &= -B_{t-1}^{-1}\lambda\mu + B_{t-1}^{-1}S_{t}&\\
    &= B_{t-1}^{-1}(-\lambda\mu + S_{t})&\\
\end{flalign*}
\vspace{-30pt}
\begin{flalign*}
    &B_{t-1}(\hat{\mu}_{t-1} - \mu) = -\lambda\mu + S_{t}&\\
    &(\hat{\mu}_{t-1} - \mu)^{T}B_{t-1}(\hat{\mu}_{t-1} - \mu)  =(\hat{\mu}_{t-1} - \mu)^{T}(-\lambda\mu + S_{t})&
\end{flalign*}
Replacing $\hat\mu - \mu$ to the right with $B_{t-1}^{-1}(-\lambda\mu + S_{t})$ into the obtained result, and then taking square roots on both sides allows us to proceed to: 
\begin{flalign*}
    &\sqrt{(\hat{\mu}_{t-1} - \mu)^{T}B_{t-1}(\hat{\mu}_{t-1} - \mu)}&\\
    &=\sqrt{(B_{t-1}^{-1}(-\lambda\mu + S_{t}))^{T}(-\lambda\mu + S_{t})}&\\
    &=\sqrt{(-\lambda\mu + S_{t})^{T}(B_{t-1}^{-1})^{T}(-\lambda\mu + S_{t})}&\\
    &=\sqrt{(-\lambda\mu + S_{t})^{T}B_{t-1}^{-1}(-\lambda\mu + S_{t})}&\\
\end{flalign*}
\vspace{-25pt}
\begin{flalign*}
    \implies&\sqrt{(\hat{\mu}_{t-1} - \mu)^{T}B_{t-1}(\hat{\mu}_{t-1} - \mu)}&\\
    &=\sqrt{(-\lambda\mu + S_{t})^{T}B_{t-1}^{-1}(-\lambda\mu + S_{t})}&\\
\end{flalign*}
\vspace{-25pt}
\begin{flalign*}
    &\implies\Vert\hat{\mu}_{t-1} - \mu \Vert_{B_{t-1}} = \Vert-\lambda\mu + S_{t} \Vert_{B_{t-1}^{-1}}&
\end{flalign*}
Using triangle inequality leads us to:
\begin{flalign*}
    \Vert\hat{\mu}_{t-1} - \mu \Vert_{B_{t-1}} \leq \Vert-\lambda\mu\Vert_{B_{t-1}^{-1}} + \Vert S_{t} \Vert_{B_{t-1}^{-1}}
\end{flalign*}

Since $B_{t-1}\succeq\lambda I$, $\Vert-\lambda\mu\Vert_{B_{t-1}^{-1}} \leq \sqrt{\lambda}$. \cite{krishnamurthy2018semiparametric} derived a tight, high-probability upper bound for $||S_{t}||_{B_{t-1}^{-1}}$ which is presented in \textbf{Lemma 2} below. \qed\\

\textbf{Lemma 2}. (\textbf{Lemma 10} of \cite{krishnamurthy2018semiparametric}) With probability $1-\delta, $  $||S_{t}||_{B_{t-1}^{-1}} \leq 3\sqrt{d \log(1 + T /(d\lambda))+ 18 \log(T /\delta)}$ for all $t\in[T]$.

The term $X_{t}$ is a vector-valued martingale difference independent of $\zeta_{t}$ given $\mathcal{F}_{t-1}$. Hence, bounding $||S_{t}||_{B_{t-1}^{-1}}$ calls for concentration inequalities for self-normalized vector-valued martingales of \cite{DELAPENA20094210}. The proof of \textbf{Lemma 2} is presented in the appendix. Due to \textbf{Lemma 2}, the following corollary is straightforward.\\

\textbf{Corollary 4}.The complexity of $\gamma(T)$ is $\mathcal{O}(\sqrt{d}\log{T})$ with a probability of at least $1-\delta$ for all $t \in [T]$.

Proof. Results of \textbf{Lemma 1} lead us to the following inequality that holds with probability at least $1-\delta$ for all $t \in [T]$ :
\begin{flalign*}
&||\hat{\mu}_{t-1} - \mu||_{B_{t-1}} \leq \Vert-\lambda\mu\Vert_{B_{t-1}^{-1}} + \Vert S_{t} \Vert_{B_{t-1}^{-1}}&\\
&\leq \sqrt{\lambda} + 3\sqrt{d\log(1 + T /(d\lambda))+ 18 \log(T /\delta)}&\\
&\leq\sqrt{\lambda} + 3\sqrt{d}\sqrt{\log(\frac{(d\lambda+T)T^{18}}{d\lambda\delta^{18}})}&
\end{flalign*}
Since complexity of $\sqrt{\lambda}$ is $\mathcal{O}(1)$, the overall complexity of the upper bound comes to $\mathcal{O}(\sqrt{d\log(T}))$. Hence,
\begin{flalign*}
    \Vert\hat{\mu}_{t-1} - \mu \Vert_{B_{t-1}} & \leq\Vert-\lambda\mu\Vert_{B_{t-1}^{-1}} + \Vert S_{t} \Vert_{B_{t-1}^{-1}} &\\
    & = \mathcal{O}(\sqrt{d\log(T})) && \qed 
\end{flalign*}

The following corollary is also straightforward from \textbf{Lemma 2}.

\textbf{Corollary 5.} \label{corrolary_5} For all $i,j \in [N]$ and all $t \in [T]$, the inequality, $(\hat{\mu}_{t-1}-\mu)^{T}(b_{t,i}-b_{t,j}) \leq \gamma(T) ||b_{t,i}-b_{t,j}||_{B_{t-1}^{-1}} $ holds with probability at least $1 - \delta$.\\

Proof. Using Cauchy-Schwarz inequality we get: \\
$(\hat{\mu}_{t-1}-\mu)^{T}(b_{t,i}-b_{t,j}) \leq ||\hat{\mu}_{t-1}-\mu||_{B_{t-1}} ||b_{t,i}-b_{t,j}||_{B_{t-1}^{-1}}.$ 
Since \textbf{Lemma 1}, binds the term $||\hat{\mu}_{t-1}-\mu||_{B_{t-1}}$ as $||\hat{\mu}_{t-1}-\mu||_{B_{t-1}} \leq \gamma(T)$, with probability of at least $1-\delta$, the following inequality hold with probability of at least $1-\delta$:\\
$(\hat{\mu}_{t-1}-\mu)^{T}(b_{t,i}-b_{t,j}) \leq \gamma(T) ||b_{t,i}-b_{t,j}||_{B_{t-1}^{-1}}$.

{\textbf{Corollary 5} guarantees that the optimal action $a_t^*$ is always included in the filtered set $\mathcal{A}_t$ with high probability.}\qed\\

\textbf{Lemma 6}. \label{lamma_6} (\textbf{Lemma 6} of \cite{krishnamurthy2018semiparametric}) With probability at least $1-\delta$, the optimal action $a^{*}_{t}$ is included in surviving action set $\mathcal{A}_{t}$ after the first filtering step for all $t \in [T]$.

Proof. The proof follows by simply setting $b_{t, i} = b_{t, a^{*}}$ in (3) of \textbf{Algorithm \ref{algo}}. For $\forall j \in [N]$,
\begin{align*}
\hat{\mu}_{t-1}(b_{t,j}-b_{t,a^{*}}) &= (\hat{\mu}_{t-1}-\mu)^{T}(b_{t,j}-b_{t,a^{*}})\\
&~~~~+ (\mu)^{T}(b_{t,j}-b_{t,a^{*}})\\
&\leq \gamma(T)||b_{t,j}-b_{t,a^{*}}||_{B_{t-1}^{-1}} + 0. 
\end{align*}
The first term is bounded due to \textbf{Corollary 5}, and second term is bounded since $a_{t}^{*} = \underset{i}{\text{argmax}}$ $ b_{t,i}^{T}\mu$, which makes the $(\mu)^{T}(b_{t,j}-b_{t,a^{*}}) \leq 0$ . Therefore, $a^{*}_{t} \in \mathcal{A}_{t}$.\qed\\

{We next prove that our proposed probability distribution function $\pi_t$ over the actions that survived the first filtering step satisfies the constraint $||b_{t,i}-\bar{b_{t}}||_{B_{t-1}^{-1}}^{2}\leq\sum_{j\in A_{t}}{\pi_j\left(t\right)||b_{j,i}-\bar{b_{t}}||_{B_{t-1}^{-1}}^2}$ given in \cite{krishnamurthy2018semiparametric} up to a multiplicative constant 4.}

\textbf{Proposition 7} Let $\pi_i(t)$ be the distribution shown in Line 8 of \textbf{Algorithm \ref{algo}} Then we have, for every $i\in \mathcal{A}_t$,
$$||b_{t,i}-\bar{b_{t}}||_{B_{t-1}^{-1}}^{2}\leq 4\sum_{j\in \mathcal{A}_{t}}{\pi_j\left(t\right)||b_{t,j}-\bar{b_{t}}||_{B_{t-1}^{-1}}^2}.$$
Proof. Since $\pi_j(t)$ is $1/2$ for $k$ and $l$ that maximizes that Mahalanobis distance $||b_{t,a} - b_{t,b}||_{B_{t-1}^{-1}}$, where $a,b \in \mathcal{A}_{t}$, and 0 for any other action,
\begin{flalign}
    &\underset{j \in \mathcal{A}_t}{\sum } {\pi_j (t)} || b_{t, j} - \Bar{b}_t||_{B_{t-1}^{-1}}^2\nonumber&\\ 
    &=\frac{1}{2} || b_{t, k} - \bar{b}_t ||_{B_{t-1}^{-1}}^2  + \frac{1}{2} || b_{t, l} - \bar{b}_t ||_{B_{t-1}^{-1}}^2\nonumber&\\
    &= \frac{1}{2} ||\frac{1}{2}(b_{t,k}-b_{t,l})||_{B_{t-1}^{-1}}^{2}+ \frac{1}{2} ||\frac{1}{2}(b_{t,k}-b_{t,l})||_{B_{t-1}^{-1}}^{2}\nonumber&\\
    &= ||\frac{1}{2}(b_{t,k}-b_{t,l})||_{B_{t-1}^{-1}}^{2} =  \frac{1}{4}||(b_{t,k}-b_{t,l})||_{B_{t-1}^{-1}}^{2}\label{eqnecessary}
\end{flalign}
Now we proceed to express the term $||b_{t,i}-\bar{b_{t}}||_{B_{t-1}^{-1}}^{2}$:
\begin{flalign*}
    \forall i \in \mathcal{A}_{t},||b_{t,i} - \bar{b_t}||_{B_{t-1}^{-1}}^{2} = &||b_{t,i} - \frac{1}{2}(b_{t,k}+b_{t,l})||_{B_{t-1}^{-1}}^{2}&
\end{flalign*}
Applying triangle inequality to the term at the right lets us write:
\begin{flalign*}
    ||b_{t,i} - \bar{b_{t}}||_{B_{t-1}^{-1}}^{2}\leq &(||\frac{1}{2}(b_{t,i}-b_{t,k})||_{B_{t-1}^{-1}}&\\
    &+||\frac{1}{2}(b_{t,i}-b_{t,l})||_{B_{t-1}^{-1}})^{2}&
\end{flalign*}
Since arm $k$ and arm $l$ maximizes that Mahalanobis distance $||b_{t,a} - b_{t,b}||_{B_{t-1}^{-1}}$, where $a,b \in \mathcal{A}_{t}$, we can bind the left term as:
\begin{flalign}
    ||b_{t,i} - \bar{b_{t}}||_{B_{t-1}^{-1}}^{2}\leq &(\frac{1}{2}||(b_{t,l}-b_{t,k})||_{B_{t-1}^{-1}}\nonumber&\\
    &+\frac{1}{2}||(b_{t,l}-b_{t,k})||_{B_{t-1}^{-1}})^2\nonumber&\\
    =&||b_{t,l}-b_{t,k}||_{B_{t-1}^{-1}}^{2}\label{lastref}&
\end{flalign}
Hence,  $ ||b_{t,i} - \bar{b_t}||_{B_{t-1}^{-1}}^{2} \leq 4 \underset{j\in \mathcal{A}_t}{\sum}{\pi_j\left(t\right)||b_{j,i}-\bar{b_{t}}||_{B_{t-1}^{-1}}^2} \qed$\\

%

{\textbf{Proposition 7} plays a crucial rule in proving the final regret bound of our algorithm. The final regret bound is presented in \textbf{Theorem 8} below, which has the same order as the one given by \cite{krishnamurthy2018semiparametric}. This is because we use a probability distribution function that satisfies the constraint given by \cite{krishnamurthy2018semiparametric} up to multiplicative constant which does not affect the order of the final bound. The proof follows the lines of the proof in \cite{krishnamurthy2018semiparametric}. }

\textbf{Theorem 8:} The complexity of upper bound on cumulative regret is $O(d\sqrt{T}\log{T})$ with probability of at least $1-\delta$.

Proof. The regret can be decomposed as:
\begin{flalign*}
    &\text{regret($t$)} =(b_{t,a_{t}^{*}}-b_{t,a_{t}})^{T}\mu&\\
    &=(b_{t,a_{t}^{*}}-b_{t,a_{t}})^{T}(\mu-\hat{\mu}_{t-1})+(b_{t,a_{t}^{*}}-b_{t,a_{t}})^{T}\hat{\mu}_{t-1}&\\  
    &\leq \gamma(T)||b_{t,a_{t}^{*}}-b_{t,a_{t}}||_{B_{t-1}^{-1}}+\gamma(T)||b_{t,a_{t}^{*}}-b_{t,a_{t}}||_{B_{t-1}^{-1}}&
\end{flalign*}
The first term in the upper bound holds as per \textbf{Corollary 5}. The second term in the upper bound holds because of the fact that $a_{t} \in \mathcal{A}_{t}$. 
Hence, the regret can be expressed as:
\begin{flalign*}
    \text{regret($t$)} &\leq2\gamma(T)||b_{t,a_{t}^{*}}-b_{t,a_{t}}||_{B_{t-1}^{-1}}
\end{flalign*}
By applying triangle inequality the expression can be bounded as:
\begin{flalign*}
    \text{regret($t$)} &\leq2\gamma(T)(||b_{t,a_{t}}-\bar{b_{t}}||_{B_{t-1}^{-1}} + ||b_{t,a_{t}^{*}}-\bar{b_{t}}||_{B_{t-1}^{-1}})
\end{flalign*}
Due to \textbf{Lemma 6}, we have $a_t^*\in\mathcal{A}_t$ with high probability. Therefore, due to (\ref{lastref}) in the proof of \textbf{Proposition 7}, we have 
\begin{flalign*}
    \text{regret($t$)} &\leq2\gamma(T)(||b_{t,k}-b_{t,l}||_{B_{t-1}^{-1}} +&\\
    &||b_{t,a_{t}}-\frac{1}{2}(b_{t,k}+b_{t,l})||_{B_{t-1}^{-1}})&
\end{flalign*}
Using the fact that selected action $a_{t} \in \{b_{t,k},{b_{t,l}}\}$ for the second term, we arrive at:
\begin{flalign*}
    \text{regret($t$)} &\leq2\gamma(T)(||b_{t,k}-b_{t,l}||_{B_{t-1}^{-1}}  &\\   
    &+ \frac{1}{2}||b_{t,k}-b_{t,l}||_{B_{t-1}^{-1}})&\\
    &\leq2\gamma(T)(\frac{3}{2}||b_{t,k}-b_{t,l}||_{B_{t-1}^{-1}}) &
\end{flalign*}
Due to (\ref{eqnecessary}), we get the following inequality:
\begin{flalign*}
    \text{regret($t$)}
    &\leq 2\gamma(T)\cdot\frac{3}{2}\sqrt{4\sum_{f\in \mathcal{A}_{t}}\pi_{f}(t)||b_{t,f}-\bar{b_{t}}||^2_{B_{t-1}^{-1}}}&\\
    &=6\gamma(T)\cdot \sqrt{E[||b_{t,a_{t}}-\bar{b_{t}}||^2_{B_{t-1}^{-1}}|\mathcal{F}_{t-1}]}.&
\end{flalign*}

By Jensen's inequality,
\begin{align*}
\sum_{t=1}^T\text{regret($t$)}\leq 6\gamma(T)\cdot\sqrt{T\sum_{t=1}^TE[||b_{t,a_{t}}-\bar{b_{t}}||^2_{B_{t-1}^{-1}}|\mathcal{F}_{t-1}]}
\end{align*}

Direct application of the elliptical potential lemma of \cite{abbasi2011improved} leads us to:
$\sum \text{regret} \leq 12\gamma(T) O(\sqrt{d T\log{T}}
) = O(d\sqrt{T}\log{T})$\qed\\

\section{EXPERIMENTS}
\begin{figure*}[ht!]
\centering
\includegraphics[width=\textwidth]{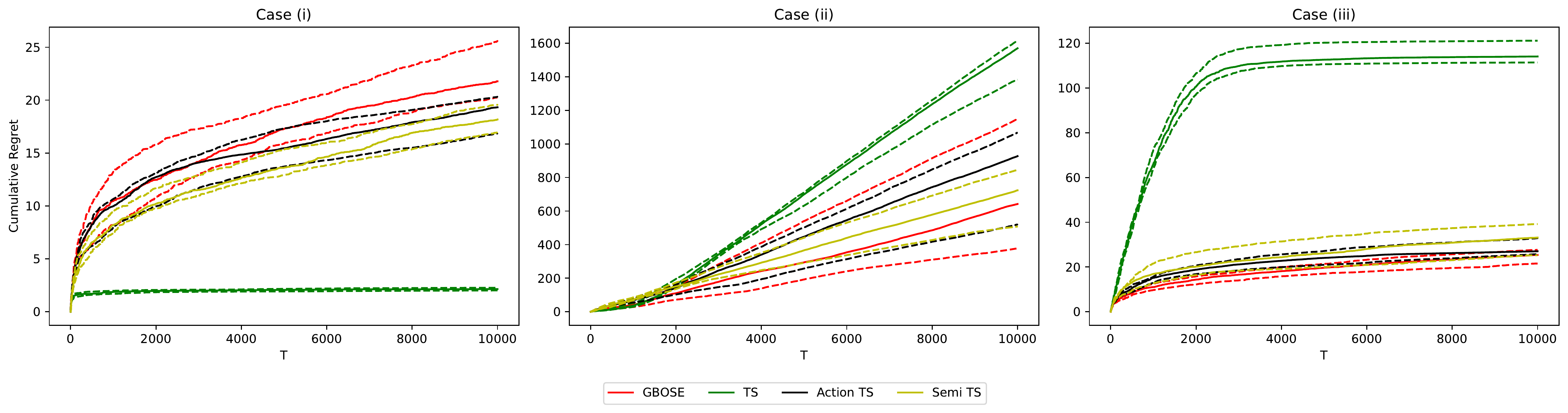}
\vspace{-5pt}
\caption{Median (solid), 1st and 3rd quartiles (dashed) of cumulative regret over 10 simulations when $N = 2$ and $d = 10$.}
\vspace{-5pt}
\label{fig:2_10}
\end{figure*}
\begin{figure*}[ht!]
\centering
\includegraphics[width=\textwidth]{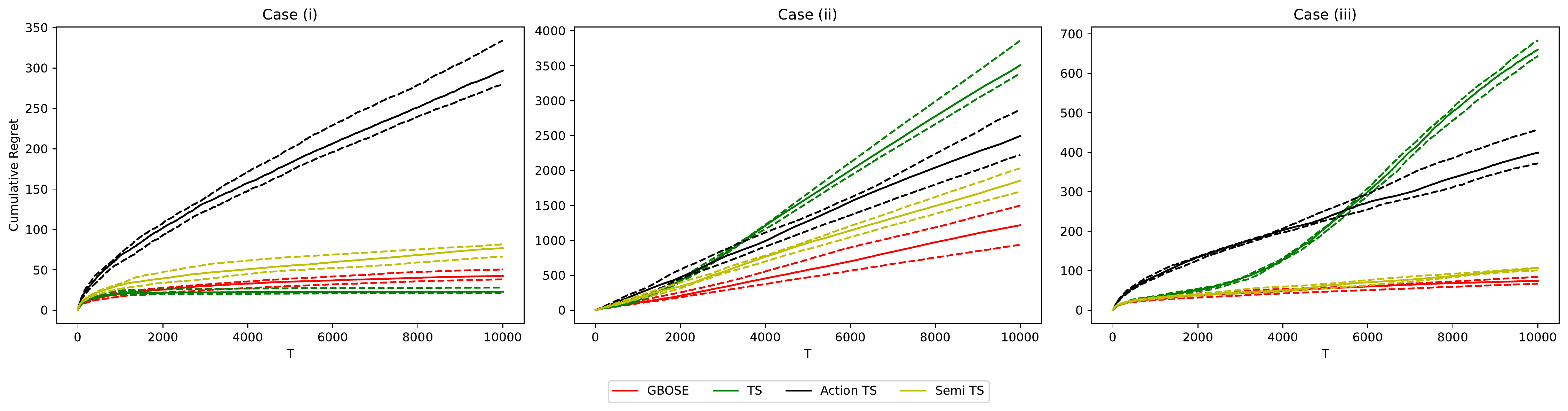}
\vspace{-5pt}
\caption{Median (solid), 1st and 3rd quartiles (dashed) of cumulative regret over 10 simulations when $N = 10$ and $d = 2$.}
\vspace{-5pt}
\label{fig:10_2}
\end{figure*}
\begin{figure*}[ht!]
\centering
\includegraphics[width=\textwidth]{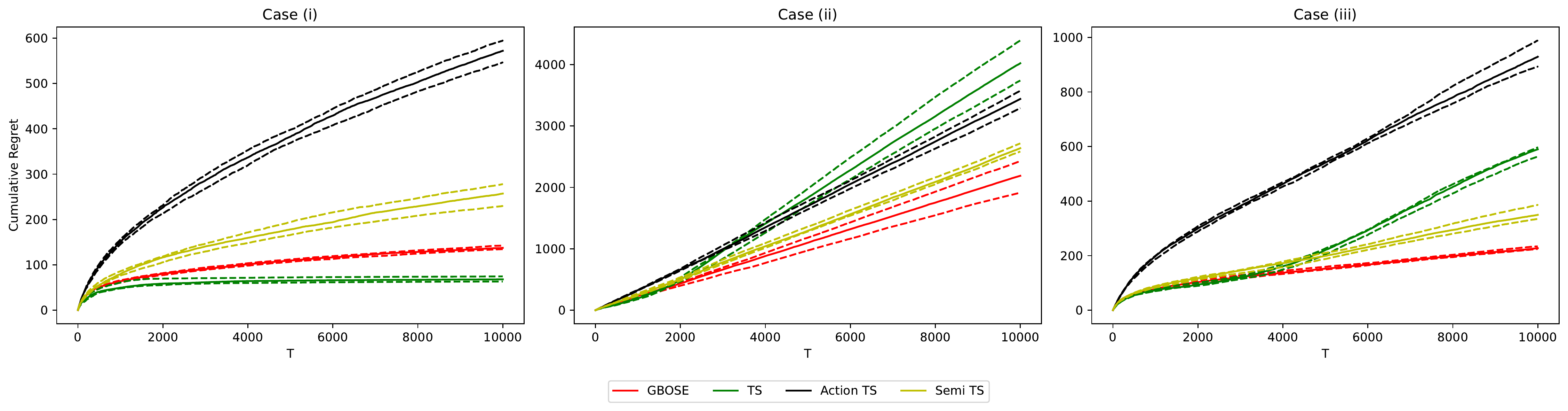}
\vspace{-5pt}
\caption{Median (solid), 1st and 3rd quartiles (dashed) of cumulative regret over 10 simulations when $N = 10$ and $d = 10$.}
\vspace{-5pt}
\label{fig:10_10}
\end{figure*}
\begin{table}[ht]
\begin{center}
\resizebox{\linewidth}{!}{
\begin{tabular}{lcclll}
\hline
\textbf{Algorithms} & $N$ & $d$  & (i) & (ii) & (iii) \\ \hline
GBOSE / BOSE   & \multirow{4}{*}{2} & \multirow{4}{*}{10}     & 21.8 & \textbf{642.0}  & \textbf{25.4}   \\
Action TS  &   &  & 19.3 & 927.1  & 27.0   \\
Semi TS    &   &  & 18.2 & 723.4  & 33.1   \\ 
TS    &    &   & \textbf{2.1} & 1569.6  & 114.1   \\ \hline
GBOSE       & \multirow{4}{*}{10}   & \multirow{4}{*}{2}   & 42.5 & \textbf{1217.5}  & \textbf{74.5}   \\
Action TS  &   & & 296.7 & 2495.4  & 399.1   \\
Semi TS    & &   & 76.9 & 1856.4  & 106.7   \\
TS    &    &   & \textbf{23.0} & 3508.8  & 659.9   \\ \hline
GBOSE       &  \multirow{4}{*}{10} &  \multirow{4}{*}{10}  & 137.6 & \textbf{2188.9}  & \textbf{227.4}   \\
Action TS  &    &  & 572.0 & 3439.6  & 928.7   \\
Semi TS    &  &   & 257.3 & 2638.0  & 349.6   \\ 
TS    &    &  & \textbf{68.1} & 4020.8  & 589.8   \\ \hline
\end{tabular}
}
\end{center}

\caption{Median of $\mathcal{R}(T)$ over 10 simulations}
\label{tab:median}
\end{table}

To evaluate our algorithm, we conduct simulation studies and compare ours with four alternative algorithms i.e., the original TS \citep{agrawal2013thompson}, action-centered TS \citep{greenewald2017action}, BOSE \citep{krishnamurthy2018semiparametric}, semiparametric TS \citep{kim2019contextual}. We consider three ($N$, $d$) setup for the experiment where $N = 2, 10, 10$ and $d = 10, 2, 10$ respectively. {We closely follow the experimental setups in \cite{kim2019contextual}.} We define context vectors for each action at timestep $t$ as $0_d$ for the first one and $b_{t,i}= [I(i=2)z_{t,i}^T,\dots,I(i=N)z_{t,i}^T]^T$ for the rest ones where $z_{t,i}$ is generated uniformly at random from $d/(N-1)$-dimensional unit sphere. Then we generate the errors from $\mathcal{N}(0,0.12)$ and the rewards from (\ref{eq2}), where we generate $\mu$ from $\mathcal{U}(-0.5,0.5)$. For each setup we consider three functions for $v(t)$: (i) $v(t)=0$, (ii) $v(t)=\log_2{(t + 1)}\sin(0.0005t)^2 + t^{1/4}$, (iii) $v(t)=-\cos(0.0005t)\sqrt{\lvert b_{t,a_{t}^{*}}^T\mu\rvert}$. Then we perform 10 iterations for each ($N$, $d$, $v(t)$) setup  under 11 different  $\gamma(t)$ values for each algorithm in order to find the best parameter that happens to have the least median cumulative regret $\mathcal{R}(t)$ among 10 repetitions.\\
\begin{flalign*}
    \mathcal{R}(t) = \sum_{\tau=1}^{t} \text{regret}(\tau)
\end{flalign*}
Figures \ref{fig:2_10}, \ref{fig:10_2}, and \ref{fig:10_10} demonstrates the change in cumulative regret $\mathcal{R}(t)$ according to the timestep $t$ with best $\gamma(t)$ values for each algorithm. Solid lines stand for the median and the dashed lines stand for 1st and 3rd percentiles of the cumulative regret. As can be seen from the figures, the original TS algorithm performs the best in linear environment (i.e., $v(t) = 0$). However, it struggles to converge when $v(t)$ becomes nonstationary function that changes with time. Semi TS seems to be the closest to us in terms of change in cumulative regret while Action TS fails to learn in many cases.  Table \ref{tab:median} summarizes the minimum regret values of each algorithm for each setup. The BOSE algorithm has no explicit method for $N > 2$, hence the results for those cases are not shown in the table. For $N=2$ both BOSE and GBOSE boil down to the same operations, so their results are same. As can be seen from the table, our algorithm achieves the lowest cumulative regret for each setup for all non-linear environments and second lowest cumulative regret for linear settings after original TS for $N > 2$. We stress once again that in reality, linear settings are uncommon, and such assumptions are often impossible to validate. Hence, we believe that trading off a modest loss in performance in the specialized linear situation for much greater resilience, as seen by our results, is preferable.

\section{Conclusion}
In this work, we proposed a new contextual MAB 
algorithm for a semi-parametric reward model, better suited
for the real-world cases. We construct specific arm selection distribution for scenarios that involve over two arms while preserving the SOTA complexity of upper bound on regret among existing semi-parametric algorithms and the simulation results confirms the superiority of our method.

\bibliography{references}
\end{document}


%

%

\onecolumn
\aistatstitle{Supplementary Materials}

\section{PROOFS}

This material includes detailed proof of \textbf{Lemma 2} skipped in the main paper and elaborates further on derivation of the probabilistic upper bound of $||\hat{\mu}_{t} - \mu||_{B}$. We show that our action selection distribution $\pi_t$ induces a {\it symmetric} martingale, which can be bounded using Lemma 10 of \cite{krishnamurthy2018semiparametric}. We present a proof that closely follows the lines of \cite{krishnamurthy2018semiparametric}.


 We observe that in $S_t$, $\{(X_{t},\zeta_{t})\}_{t=1}^{T}$ is a stochastic process with $X_{t} \in \mathbb{R}^{d}$ and $\zeta_{t} \in \mathbb{R}$ such that (\rom{1}) $(X_{t},\zeta_{t})$ is $\mathcal{F}_{t}$ measurable, (\rom{2}) $|\zeta_{t} \leq 3|$ for all $t\in[T]$, (\rom{3}) $X_{t} \perp\!\!\!\!\perp \zeta_{t}|\mathcal{F}_{t-1}$, (\rom{4}) $\mathbb{E}[X_{t}|\mathcal{F}_{t}] = 0$, and, (\rom{5}) for all $\lambda \in \mathbb{R}^{d}$, $\mathcal{L}(\langle \lambda ,X_{t}|\mathcal{F}_{t}\rangle) = \mathcal{L}(-\langle \lambda ,X_{t}|\mathcal{F}_{t}\rangle)$  where $\mathcal{L}$ denotes the probability law, so that $X_{t}$ is conditionally symmetric. 

The fact (\rom{5}) can be easily shown as follows. We have,
$$X_t=b_{t,a_t}-\bar{b}_t=b_{t,a_t}-\frac{1}{2}\left(b_{t,k}+b_{t,l}\right),$$
where $b_{t,k}$ and $b_{t,l}$ are the two arms that maximize the Mahalanobis distance, $||b_{t,a}-b_{t,b}||_{B_{t-1}^{-1}}$. Since our distribution $\pi_t$ imposes probability $\frac{1}{2}$ to $b_{t,k}$ and $b_{t,l}$ respectively and zero probability to any other action, we also have 
$$b_{t,a_t}=\left\{\begin{array}{cr}b_{t,k}& \text{with probability } \frac{1}{2}\\b_{t,l}& \text{with probability } \frac{1}{2}\end{array}\right.$$
Therefore, we have 
$$X_t=\left\{\begin{array}{cr}\frac{1}{2}\left(b_{t,k}-b_{t,l}\right)& \text{with probability } \frac{1}{2}\\\frac{1}{2}\left(b_{t,l}-b_{t,k}\right)& \text{with probability } \frac{1}{2}\end{array}\right.$$
Hence, $X_t$ and $-X_t$ have identical distributions, implying that $X_t$ is a symmetric martingale difference sequence. We can now directly apply Lemma 10 of  \cite{krishnamurthy2018semiparametric}.

In the case when only one action survives the filtering step, we simply assign probability 1 to the unique surviving action and pull it. In this case, $X_t=0$ which is also symmetric martingale difference sequence. Due to Lemma 6, the unique survivor action will be the optimal action $a_t^*$ wih high probability, so we incur no regret at that round.

\textbf{Lemma 10} of \cite{krishnamurthy2018semiparametric}. Let $\{\mathcal{F}_{t}\}$ be a filtration and let $\{(X_{t},\zeta_{t})\}_{t=1}^{T}$ be a stochastic process with $X_{t} \in \mathbb{R}^{d}$ and $\zeta_{t} \in \mathbb{R}$ such that (\rom{1}) $(X_{t},\zeta_{t})$ is $\mathcal{F}_{t}$ measurable, (\rom{2}) $|\zeta_{t} \leq C|$ for all $t\in[T]$, (\rom{3}) $X_{t} \perp\!\!\!\!\perp \zeta_{t}|\mathcal{F}_{t}$, (\rom{4}) $\mathbb{E}[X_{t}|\mathcal{F}_{t}] = 0$, and, (\rom{5}) for all $\lambda \in \mathbb{R}^{d}$, $\mathcal{L}(\langle \lambda ,X_{t}|\mathcal{F}_{t}\rangle) = \mathcal{L}(-\langle \lambda ,X_{t}|\mathcal{F}_{t}\rangle)$ where $\mathcal{L}$ denotes the probability law, so that $X_{t}$ is conditionally symmetric. Let $\Sigma \triangleq X_{t}X_{t}^{T}$. Then for any positive definite matrix $Q$ we have,

\begin{align*}
    \mathbb{P}\left[\left\Vert \sum_{t=1}^T X_t\zeta_t\right\Vert_{(Q+C^2\Sigma)^{-1}}^2\geq 2\log\left(\frac{1}{\delta} \sqrt{\frac{\det(Q+C^2\Sigma)}{\det(Q)}}\right)\right] \leq \delta
\end{align*}

\textbf{Lemma 10} of \cite{krishnamurthy2018semiparametric} allows us to write, that with probability of at least $1-\delta$ :
\begin{align*}
\Vert\sum_{\tau =1}^{t-1}X_{\tau}\zeta_{\tau}\Vert_{B_{t-1}^{-1}} = C^{2}\Vert\sum_{\tau =1}^{t-1}X_{\tau}\zeta_{\tau}\Vert_{C^{2}B_{t-1}^{-1}}^{2}
\leq2C^2\log\left(\frac{1}{\delta}\sqrt{\frac{\det(C^{2}B_{t-1})}{\det(C^{2}\lambda I)}}\right)
\end{align*}
This inequality comes from \textbf{Lemma 10} of \cite{krishnamurthy2018semiparametric} after setting $Q = C^{2}\lambda I$. Then using  the relationship $\det(aQ)=a^{d}\det(Q)$, that holds for our $d\times d$ positive definite matrix $Q$, we move to the next step:
\begin{align*}
\Vert\sum_{\tau =1}^{t-1}X_{\tau}\zeta_{\tau}\Vert_{B_{t-1}^{-1}} \ \leq2C^2\log\left(\frac{1}{\delta}\sqrt{\frac{C^{2d}\det(B_{t-1})}{C^{2d}\det(\lambda I)}}\right)=2C^2\log\left(\frac{1}{\delta}\sqrt{\frac{\det(B_{t-1})}{\det(\lambda I)}}\right)
=2C^2\log\left(\frac{1}{\delta}\sqrt{\frac{(\det(B_{t-1})^{1/d})^d}{\lambda^d}}\right)
\end{align*}


\textbf{Lemma 3.} (\textbf{Lemma 9} in \cite{krishnamurthy2018semiparametric}) Let $X_{i}\in\mathbb{R}^{d}$ for $i\in[n]$ and let $\Vert X_{i}\Vert_{2} \leq L$. Define $B \triangleq \lambda I + \sum_{i=1}^{n}X_{i}X_{i}^{T}$. Then $\det(B) \leq (\lambda + nL^{2}/d)$.

\begin{equation*}
\det(B)^{\frac{1}{d}} \leq \frac{1}{d}\text{tr}(B)
                    =\frac{1}{d}\text{tr}(\lambda I) + \frac{1}{d}\text{tr}\left(\sum_{i=1}^{n}X_{i}X_{i}^{T}\right)
                     =\frac{1}{d}\text{tr}(\lambda I) + \frac{1}{d}\text{tr}\left(\sum_{i=1}^{n} \Vert X_{i} \Vert_{2}^{2}\right)
                     \leq \lambda + \frac{nL^2}{d}  \qed 
\end{equation*}

By applying \textbf{Lemma 3} on the derived expression, we can get the following inequality:
$$||\sum_{\tau =1}^{t-1}X_{\tau}\zeta_{\tau}||_{B_{t-1}^{-1}}\leq 2C^2\log\left(\frac{1}{\delta}\sqrt{\frac{(\lambda + \frac{T}{d})^d}{\lambda^d}}\right)=2C^2\log\left(\frac{1}{\delta}\sqrt{\left(1 + \frac{T}{\lambda d}\right)^d}\right)$$
Using the definition $|\zeta_{\tau}|\leq3 \triangleq C$, we can proceed:
$$||\sum_{\tau =1}^{t-1}X_{\tau}\zeta_{\tau}||_{B_{t-1}^{-1}}\leq2(3)^2\log\left(\frac{1}{\delta}\sqrt{\left(1 + \frac{T}{\lambda d}\right)^d}\right)
=18\log\left(\frac{1}{\delta}\sqrt{\left(1 + \frac{T}{\lambda d}\right)^d}\right)=9d\log\left(1 + \frac{T}{\lambda d}\right)+18\log\left(\frac{1}{\delta}\right)$$\\

Taking a union bound over all rounds $T$ leads us to:

$$||\hat{\mu}_{t-1} - \mu||_{B_{t-1}} \leq\sqrt{\lambda}+||\sum_{\tau =1}^{t-1}X_{\tau}\zeta_{\tau}||_{B_{t-1}^{-1}}\leq \sqrt{\lambda}+\sqrt{9d\log\left(1 + \frac{T}{\lambda d}\right)+18\log\left(\frac{T}{\delta}\right)}\qed$$




\bibliography{references}